\documentclass[9pt, conference]{IEEEtran}
\IEEEoverridecommandlockouts
\usepackage{cite}
\usepackage{amsmath,amssymb,amsfonts}
\usepackage{algorithmic}
\usepackage{graphicx}
\usepackage{textcomp}
\usepackage{multirow}
\usepackage{booktabs}
\usepackage{graphicx}
\usepackage[table,xcdraw]{xcolor}
\usepackage{siunitx}
\usepackage{authblk}

\def\BibTeX{{\rm B\kern-.05em{\sc i\kern-.025em b}\kern-.08em
    T\kern-.1667em\lower.7ex\hbox{E}\kern-.125emX}}
\begin{document}

\title{Boosting Text-To-Image Generation via Multilingual Prompting in Large Multimodal Models
\thanks{This work was supported in part by the National Science Foundation of China (No.62276056), the Natural Science Foundation of Liaoning Province of China (2022-KF-16-01), the Fundamental Research Funds for the Central Universities (Nos. N2216016 and N2316002), the Yunnan Fundamental Research Projects (No. 202401BC070021), and the Program of Introducing Talents of Discipline to Universities, Plan 111 (No.B16009).\\
* This work is done during the internship at Northeastern University NLP Lab.\\
† Corresponding author.}
}

\author{
  Yongyu Mu\textsuperscript{1}, Hengyu Li\textsuperscript{1}, Junxin Wang\textsuperscript{2*}, Xiaoxuan Zhou\textsuperscript{1*}, Chenglong Wang\textsuperscript{1}, Yingfeng Luo\textsuperscript{1}, Qiaozhi He,\\
  Tong Xiao\textsuperscript{1,3†}, Guocheng Chen\textsuperscript{1} and Jingbo Zhu\textsuperscript{1,3}\\
  \textsuperscript{1}School of Computer Science and Engineering, Northeastern University, Shenyang, China\\
  \textsuperscript{2}School of Economics and Management, Dalian Jiaotong University, Dalian, China\\
  \textsuperscript{3}NiuTrans Research, Shenyang, China
}

\maketitle

% #################################################
% #################################################
% ##################  Abstract  ###################
% #################################################
% #################################################
\begin{abstract}
Previous work on augmenting large multimodal models (LMMs) for text-to-image (T2I) generation has focused on enriching the input space of in-context learning (ICL). This includes providing a few demonstrations and optimizing image descriptions to be more detailed and logical. However, as demand for more complex and flexible image descriptions grows, enhancing comprehension of input text within the ICL paradigm remains a critical yet underexplored area. In this work, we extend this line of research by constructing parallel multilingual prompts aimed at harnessing the multilingual capabilities of LMMs. More specifically, we translate the input text into several languages and provide the models with both the original text and the translations. Experiments on two LMMs across 3 benchmarks show that our method, PMT2I, achieves superior performance in general, compositional, and fine-grained assessments, especially in human preference alignment. Additionally, with its advantage of generating more diverse images, PMT2I significantly outperforms baseline prompts when incorporated with reranking methods. Our code and parallel multilingual data can be found at https://github.com/takagi97/PMT2I.
\end{abstract}


% #################################################
% #################################################
% ###############  Introduction  ##################
% #################################################
% #################################################
\section{Introduction}
Texts capture the boundless imagination of humanity through their free-form and open-ended nature. This characteristic presents a considerable challenge in text-to-image (T2I) generation, which strives to create images that fully embody the essence of the text. Early works along this line of research guide image synthesis models via built-in small text encoders \cite{DBLP:conf/icml/NicholDRSMMSC22,DBLP:conf/icml/RameshPGGVRCS21}. Subsequently, it is proven that scaling text encoders offers significant advantages in generating faithful and text-aligned images \cite{DBLP:conf/nips/SahariaCSLWDGLA22}, highlighting the critical role of comprehending text input. More recently, large multimodal models (LMMs), i.e., end-to-end trained multimodal generative models and multimodal systems that integrate large language models (LLMs), have emerged as a groundbreaking approach to modal fusion. With advanced text encoding and cross-modal capabilities, these models serve as a new solution for the T2I generation task \cite{DBLP:conf/acl/CaffagniCBMS0CC24}.

Prompt engineering continues to be a focus of interest because it provides a cost-effective way to enhance the performance of LMMs in T2I generation tasks. One line of research involves enriching and detailing the image description to generate more photorealistic images \cite{DBLP:conf/chi/LiuC22a}. Meanwhile, some works leverage few-shot learning to adapt LMMs to user-specific generation scenarios via a few demonstrations \cite{DBLP:journals/corr/abs-2312-13286}. Others utilize the chain-of-thought technique to optimize the input text to follow a logical and coherent flow \cite{yao2024promptcot}. As demand for more complex and flexible image descriptions grows, improving comprehension of input content by leveraging powerful in-context learning (ICL) capabilities remains a critical yet underexplored area.

\begin{figure*}
\centering
\includegraphics[width=1.0\textwidth]{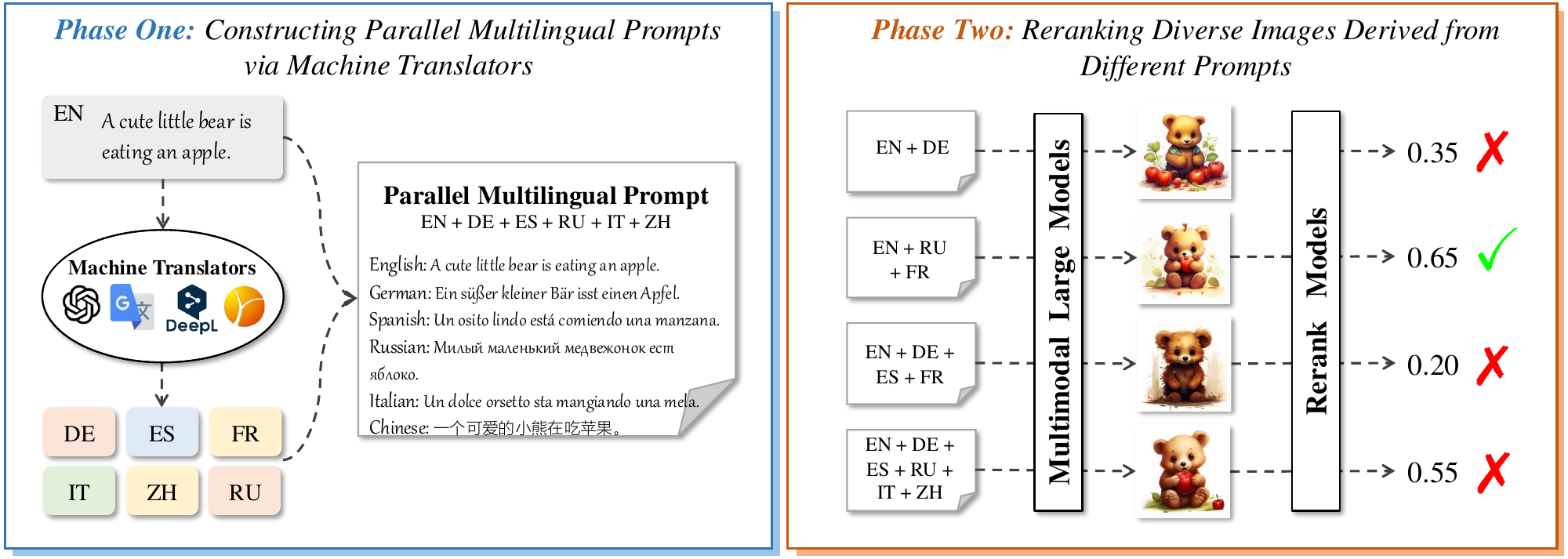}
\caption{An illustration of the two phases of PMT2I.}
\vspace{-0.9em}
\label{f1}
\end{figure*}

Recently, researchers have been aware that properly leveraging multilingualism in LLMs can boost their performance on downstream tasks \cite{DBLP:conf/emnlp/HuangTZZSXW23,DBLP:conf/emnlp/0001CWHC23,DBLP:conf/acl/ZhangLF0J0B24}. In particular, simultaneously triggering multiple language capabilities has been demonstrated to be beneficial for the models' comprehension of the input content \cite{DBLP:journals/corr/abs-2403-09073}. Considering the mixture of multilingual samples in large multimodal datasets \cite{DBLP:conf/nips/SchuhmannBVGWCC22}, we envision that LMMs also contain triggerable multilingual capabilities. In this work, we propose a \textbf{P}arallel \textbf{M}ultilingual prompting method for \textbf{T}ext-\textbf{to}-\textbf{I}mage tasks, denoted as \textbf{PMT2I}. It first translates the original image description into several languages and then triggers multiple language capabilities by providing these translations with the original text to models. On the one hand, PMT2I enriches the input space of ICL via multilingual texts. Furthermore, by varying the number of languages or altering their combinations, PMT2I can be easily scaled to generate a vast array of different prompts, thereby facilitating the generation of diverse images. When incorporated with the reranking methods, its performance can be further improved.

Extensive experimental results on two LMMs show that PMT2I achieves superior performance on general T2I synthesis. For instance, it raises the CLIP-T and CLIP-I score by $0.6$ and $0.9$ points for Emu2-Gen \cite{DBLP:journals/corr/abs-2312-13286} on the MS-COCO 30K dataset. Besides, it facilitates generating color, shape, and texture-aligned images with averaging $4.96$ points improvement across two models on the B-VQA score. Moreover, PMT2I demonstrates significant improvements in the ImageReward, suggesting that the images it generates align more closely with human preferences. When armed with the reranking method, PMT2I further surpasses baselines, demonstrating the superiority of PMT2I in terms of quality and diversity. Our contributions are summarized as follows:
\begin{itemize}
    \item We present a simple, effective, and scalable ICL method for LMMs on T2I generation tasks, with the entire process in a training-free manner.
    \item Enhancing the performance of LMMs on English prompts is a challenge since models typically comprehend English text best. However, our method, PMT2I, follows a novel weak-to-strong pattern by utilizing non-English languages to boost performance on English inputs. Additionally, PMT2I can be easily scaled to thousands of different prompts, enabling the production of a wider variety of images compared to monolingual prompts.
    \item As the multilingual capabilities of LMMs continually develop, it becomes crucial to explore multilingual ICL methods for these models. As far as we know, our work represents the first effort in this direction.
\end{itemize}

% #################################################
% #################################################
% ###############  Methodology  ###################
% #################################################
% #################################################
\section{Methodology}
This work aims to reconstruct the input for T2I tasks to enrich the input and output space of LMMs and finally generate high-quality and diverse images. We consider this a two-phase process: (1) translating the original input text to multiple languages and assembling PMT2I prompts; (2) leveraging the diversity of PMT2I prompts to generate images and reranking these results. The entire process is illustrated by Fig. \ref{f1}.

\subsection{Constructing Parallel Multilingual Prompts}
Given a piece of text $T$ describing an image $V$, LMMs project it into the latent space and then leverage its representation $C$ to guide the image generation of visual decoders. There are two mainstream choices of the visual decoder. The first one involves the denoising diffusion probabilistic model (DDPM) \cite{DBLP:conf/nips/HoJA20}, synthesizing images in an auto-regressive denoising fashion within $I$ steps, It can be shown as $P(x_I) \prod_{i=1}^I P(x_{I-i}|x_{I-i+1}, C)$, where $x_i$ denotes the intermediate result. In this way, LMMs take charge of providing conditional embedding $C$, e.g., text embedding, to DDPMs. Others utilize VQ-GANs \cite{DBLP:conf/cvpr/EsserRO21} as image de-tokenizers, which directly convert the visual tokens $C$ into pixel values.

In this process, the role of LMMs is to transform the text input $T$ into the latent representation $C$, like this:
\begin{align}
C & = \mathrm{Encode}(f(T))
\end{align}
where $f(\cdot)$ is a template that packs the input $T$ into a prompt. Since the representation $C$ is crucial for guiding the image generation, previous works enhance it by turning the input text into a more detailed and coherent one $T'$ or supplying $m$ text-image demonstrations $\{(T_1, V_1), \ldots, (T_m, V_m)\}$. Compared to them, PMT2I enriches the input space of ICL by providing different text descriptions in multiple languages to LMMs, shown as follows:
\begin{align}
C & = \mathrm{Encode}(f(S_n,T))
\end{align}
where $S_n$ denotes a set containing $n$ semantic-parallel translations in multiple languages. The template $f(\cdot)$ of PMT2I prompts is ``[\textit{language}]: [\textit{text in that language}]'', as shown in Fig. \ref{f1}. Indeed, PMT2I is inspired by the natural language processing (NLP) practice of simultaneously triggering the capabilities of multiple languages in the LLMs to enhance their comprehension \cite{DBLP:journals/corr/abs-2403-09073}. Our work extends this line of research to LMMs on multimodal tasks.
 
\begin{table*}[ht]
\LARGE
\begin{center}
\caption{Experiments on the MS-COCO, DrawBench, and three subsets of T2I-CompBench. EN Prompt stands for the original English image description. Auto-opt means the English prompt is optimized by Promptist. PMT2I-$n$ stands for utilizing $n$ translation equivalents. We rerank the image candidates of: PMT2I-1 (EN+DE), PMT2I-2 (EN+RU+FR), PMT2I-3 (EN+DE+ES+FR), PMT2I-4 (EN+DE+ES+RU+IT+ZH), PMT2I-5 (EN+ES+RU+FR+ZH+DE), and PMT2I-6 (EN+ES+RU+IT+ZH+FR+DE). Reward represents the ImageReward score. A higher value for each of these metrics indicates better performance.}
\centering
\resizebox{1.0\textwidth}{!}{
\begin{tabular}{ll|rrrr|rrr|rr|rr|rr}
\toprule
\multicolumn{2}{l|}{\multirow{2}{*}{\textbf{System}}} & \multicolumn{4}{c|}{\textbf{MS-COCO (30K)}} & \multicolumn{3}{c|}{\textbf{DrawBench (200)}} & \multicolumn{2}{c|}{\textbf{Color (300)}} & \multicolumn{2}{c|}{\textbf{Shape (300)}} & \multicolumn{2}{c}{\textbf{Texture (300)}} \\
\cmidrule(lr){3-6} \cmidrule(lr){7-9} \cmidrule(lr){10-11} \cmidrule(lr){12-13} \cmidrule(lr){14-15}
\multicolumn{2}{c|}{} & CLIP-T & CLIP-I & DINO & Reward & GPT-4o & CLIP-T & Reward & CLIP-T & B-VQA & CLIP-T & B-VQA & CLIP-T & B-VQA \\
\midrule
\multirow{6}{*}{\textbf{Emu2-Gen}} 
& EN Prompt & 29.8 & 68.8 & 45.9 & -0.205 & 38.5\% & 29.0 & -0.120 & 29.8 & 35.5\% & 28.6 & 35.9\% & 28.5 & 39.8\% \\
& + Rerank & 30.7 & 69.4 & 46.2 & -0.155 & 38.3\% & 30.5 & -0.052 & 31.1 & 36.4\% & 30.0 & 36.8\% & 30.1 & 41.4\% \\
& + Auto-opt & 26.8 & 57.8 & 36.9 & -0.464 & 31.0\% & 26.8 & -0.303 & 26.9 & 1.5\% & 26.5 & 2.0\% & 24.5 & 11.0\% \\
& \cellcolor{gray!20}PMT2I-3 & \cellcolor{gray!20} 30.4 & \cellcolor{gray!20} 69.7 & \cellcolor{gray!20} 46.2 & \cellcolor{gray!20} 0.021 & \cellcolor{gray!20}39.4\% & \cellcolor{gray!20} 29.6 & \cellcolor{gray!20}-0.035 & \cellcolor{gray!20} 30.9 & \cellcolor{gray!20}44.2\% & \cellcolor{gray!20} 28.7 & \cellcolor{gray!20}36.4\% & \cellcolor{gray!20} 29.3 & \cellcolor{gray!20}42.7\% \\
& \cellcolor{gray!20}PMT2I-6 & \cellcolor{gray!20} 30.4 & \cellcolor{gray!20} 69.5 & \cellcolor{gray!20} 46.2 & \cellcolor{gray!20} \textbf{0.075} & \cellcolor{gray!20}44.3\% & \cellcolor{gray!20} 29.6 & \cellcolor{gray!20} 0.033 & \cellcolor{gray!20} 31.1 & \cellcolor{gray!20}47.2\% & \cellcolor{gray!20} 28.7 & \cellcolor{gray!20}36.3\% & \cellcolor{gray!20} 28.9 & \cellcolor{gray!20}45.3\% \\
& \cellcolor{gray!20}+ Rerank & \cellcolor{gray!20}\textbf{31.6} & \cellcolor{gray!20}\textbf{70.6} & \cellcolor{gray!20}\textbf{46.8} & \cellcolor{gray!20} 0.005 & \cellcolor{gray!20}\textbf{45.8\%} & \cellcolor{gray!20}\textbf{31.5} & \cellcolor{gray!20} \textbf{0.231} & \cellcolor{gray!20}\textbf{33.2} & \cellcolor{gray!20}\textbf{47.8\%} & \cellcolor{gray!20}\textbf{30.8} & \cellcolor{gray!20}\textbf{40.1\%} & \cellcolor{gray!20}\textbf{31.4} & \cellcolor{gray!20}\textbf{47.9\%} \\

\midrule
\multirow{6}{*}{\textbf{Lumina-Next}} & EN Prompt & 30.8 & 67.6 & 44.2 & 0.396 & 41.9\% & 30.2 & 0.421 & 32.9 & 48.6\% & 29.8 & 33.9\% & 30.4 & 39.4\% \\
& + Rerank & 32.6 & 68.7 & 44.8 & 0.307 & 45.5\% & 32.1 & 0.592 & \textbf{35.0} & 53.8\% & 31.8 & 38.0\% & 32.8 & 47.6\% \\
& + Auto-opt & 29.5 & 61.4 & 41.3 & 0.413 & 39.0\% & 29.4 & 0.461 & 32.0 & 0.7\% & 29.0 & 2.3\% & 29.0 & 1.9\% \\
& \cellcolor{gray!20}PMT2I-3 & \cellcolor{gray!20} 31.0 & \cellcolor{gray!20} 68.3 & \cellcolor{gray!20} 46.5 & \cellcolor{gray!20}\textbf{0.529} & \cellcolor{gray!20}45.5\% & \cellcolor{gray!20} 30.4 & \cellcolor{gray!20} 0.539 & \cellcolor{gray!20} 33.1 & \cellcolor{gray!20}48.3\% & \cellcolor{gray!20} 30.1 & \cellcolor{gray!20}37.8\% & \cellcolor{gray!20} 30.4 & \cellcolor{gray!20}44.0\% \\
& \cellcolor{gray!20}PMT2I-6 & \cellcolor{gray!20} 30.9 & \cellcolor{gray!20} 68.2 & \cellcolor{gray!20} 46.7 & \cellcolor{gray!20} 0.501 & \cellcolor{gray!20}44.3\% & \cellcolor{gray!20} 30.3 & \cellcolor{gray!20} 0.504 & \cellcolor{gray!20} 33.0 & \cellcolor{gray!20}48.9\% & \cellcolor{gray!20} 30.2 & \cellcolor{gray!20}37.4\% & \cellcolor{gray!20} 30.6 & \cellcolor{gray!20}47.1\% \\
& \cellcolor{gray!20}+ Rerank & \cellcolor{gray!20}\textbf{32.7} & \cellcolor{gray!20}\textbf{69.3} & \cellcolor{gray!20}\textbf{46.9} & \cellcolor{gray!20} 0.391 & \cellcolor{gray!20}\textbf{47.8\%} & \cellcolor{gray!20}\textbf{32.2} & \cellcolor{gray!20}\textbf{0.684} & \cellcolor{gray!20}\textbf{35.0} & \cellcolor{gray!20}\textbf{55.2\%} & \cellcolor{gray!20}\textbf{32.2} & \cellcolor{gray!20}\textbf{43.3\%} & \cellcolor{gray!20}\textbf{32.9} & \cellcolor{gray!20}\textbf{50.9\%} \\

\bottomrule
\end{tabular}
}
\vspace{-0.9em}
\label{t1}
\end{center}
\end{table*}

Additionally, we outline several criteria for implementing PMT2I:
\begin{itemize}
    \item \textbf{Select languages that the model can understand.} In practice, we select several rich-resource languages based on their proportion in widely used corpora, i.e., Russian (RU), Spanish (ES), German (DE), French (FR), Chinese (ZH), and Italian (IT).
    \item \textbf{The better the translations, the better the images generated.} Since images are generated according to the original text and its translations, unfaithful translations may mislead the model, thus well-performed machine translators should be used\footnote{We used prompt-motivated GPT-4o (https://platform.openai.com), DeepL (https://www.deepl.com), and NiuTrans (https://niutrans.com) to translate.}.
    \item \textbf{Position the language that models best comprehend at the start of the sequence.} To this end, we place the English text first, as shown in Fig. \ref{f1}.
    \item \textbf{Apply PMT2I to models that incorporate multilingual capabilities.} Although models with more powerful multilingual capabilities understand PMT2I prompts better, experimental results indicate that models unintentionally trained with multilingual data can also benefit from PMT2I, e.g., Emu2-Gen \cite{DBLP:journals/corr/abs-2312-13286}.
\end{itemize}

\subsection{Scaling PMT2I and Reranking Generated Images}
Given the original text and its parallel translations in multiple languages, we can adjust the number and the arrangement of the languages to assemble diverse PMT2I prompts. Here, we theoretically analyze the maximum number of prompts derivable from this process. Assuming translations are available in $n$ languages, we have $n$ distinct sets, denoted as $\{S_1, \ldots, S_n\}$, with each set $S_i$ containing $i$ translations. Next, we do not need to consider the position of the original text, as it can be assumed to be the best-understood language and be therefore placed at the start. Thus, this results in $A^{i}_{n}$ permutations for each set. Summing up, the total number of unique language sequence permutations across all sets is given by $\sum_{i=1}^{n} A^{i}_{n}$. For instance, given $n=6$ languages, we can assemble $1956$ different PMT2I prompts. Through the above analysis, PMT2I has proven to be significantly more scalable than traditional text inputs, with the potential of generating diverse images.

Once various images are generated, an obvious next step is to rerank these candidates and choose the best image as the final output. More specifically, this process involves a rerank model that scores every image candidate according to the image description $T$. In practice, we leverage the commonly used CLIP-T scoring method \cite{DBLP:conf/emnlp/HesselHFBC21} to conduct this evaluation. It measures the similarity between the generated image and the text prompt by calculating the cosine similarity between their embedding processed by the CLIP model. The whole process can be expressed as follows:
\begin{align}
\hat{i} & = \arg \max_{i} ~ \frac{R^V_i \cdot R^T}{\|R^V_i\| \|R^T\|}
\end{align}
where $R^T$ and $R^V_i$ stand for the latent representations of the input text and the $i$-th image, respectively. The $\hat{i}$-th image is then chosen as the final output through this process.

% #################################################
% #################################################
% ################  Experiment  ###################
% #################################################
% #################################################
\section{Experiment}

\subsection{Evaluation Setups}

\paragraph{Benchmarks} We first evaluated the general T2I generation on 30k randomly sampled data from the MS-COCO \cite{DBLP:conf/eccv/LinMBHPRDZ14} validation set. Aligning with DALL-E 3 \cite{betker2023improving}, we also conducted experiments on the DrawBench \cite{DBLP:conf/nips/SahariaCSLWDGLA22} and 3 subsets of the T2I-CompBench \cite{DBLP:conf/nips/HuangSXLL23} benchmark. The former measures a system’s performance on compositional prompts while the latter offers fine-grained assessments on color, shape, and texture binding. The image descriptions of these benchmarks are in English. And we translated these English texts into six languages. For the MS-COCO dataset, we employed GPT-4o to translate into German and Spanish, DeepL for French and Italian translations, and NiuTrans for Russian and Chinese. For other benchmarks, GPT-4o was utilized to translate into all six languages.

\begin{figure*}
\centering
\includegraphics[width=1.0\textwidth]{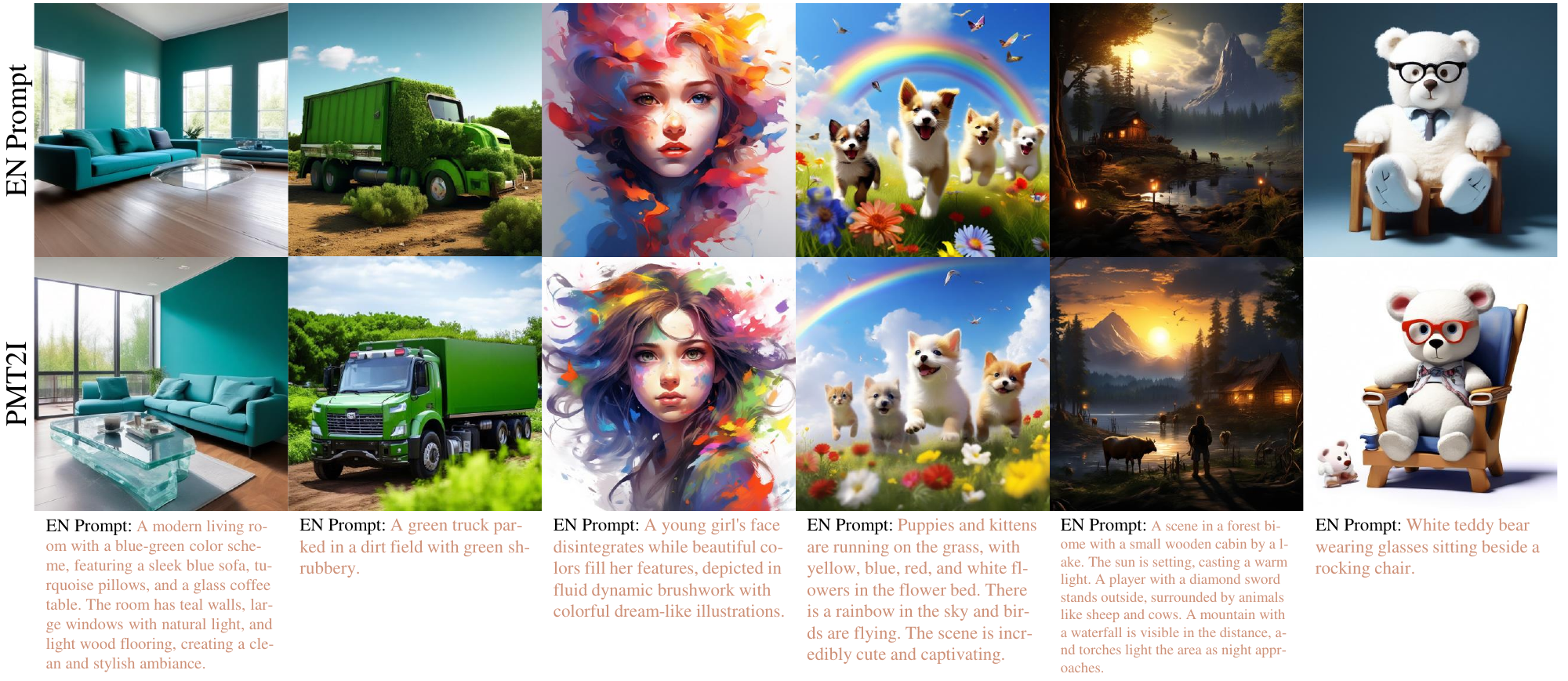}
\caption{Results of EN Prompt and PMT2I by Lumina-Next.}
\label{f2}
\end{figure*}

\begin{figure*}
\centering
\includegraphics[width=1.0\textwidth]{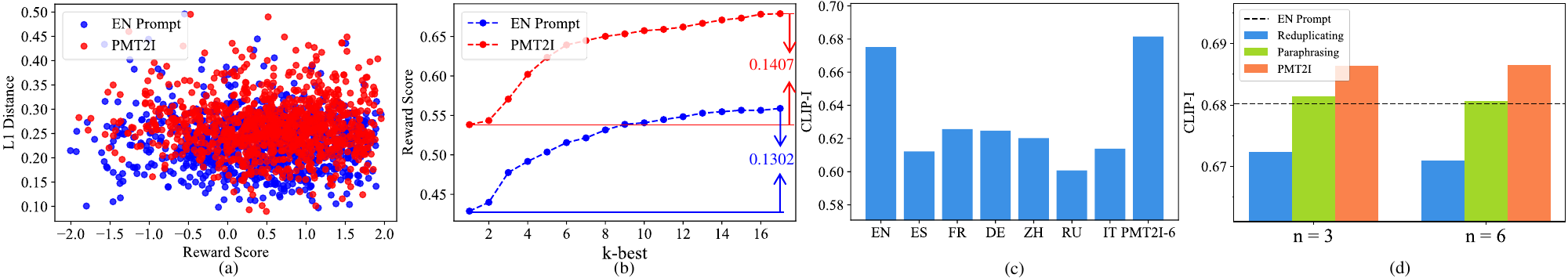}
\caption{Fig. (a) displays the L1 distances and the reward scores of 2000 randomly selected samples, where the top-right samples exhibit both high quality and diversity. Fig. (b) shows the performance of utilizing $k$ candidates for reranking. Fig. (c) presents the performance of each translation and their combination-PMT2I-6. Finally, Fig. (d) compares the performance of the baseline, reduplication, paraphrasing, and PMT2I prompts with $n$ equivalents.}
\vspace{-0.9em}
\label{f3}
\end{figure*}

\paragraph{Metrics} We used the evaluation scripts released by MagicBrush\footnote{https://github.com/OSU-NLP-Group/MagicBrush/tree/main/evaluation} \cite{DBLP:conf/nips/ZhangMCSS23}, which calculated the CLIP-T score to assess the text-image alignment and the CLIP-I and DINO scores to measure image similarity\footnote{The CLIP model we used is ViT-B/32, and the DINO model is dino\_vits16.}. A higher score means the generated image is more similar to the prompt or the real image. Additionally, for the DrawBench benchmark, we utilized GPT-4o to judge whether the generated image follows the input text (i.e., ``correct''/``incorrect'') and calculated the correct proportion. The prompt used for the judgment remains the same as the one employed by DALL-E 3. Moreover, following T2I-CompBench, we used the disentangled BLIP-VQA model \cite{DBLP:conf/icml/0001LXH22} to evaluate the results on this benchmark. Indeed, scoring by GPT-4o and BLIP-VQA offers a reverse evaluation via visual question answering. Furthermore, to assess whether the generated images align with human preferences, we utilized ImageReward \cite{DBLP:conf/nips/XuLWTLDTD23} to provide quantitative scoring.
% Since the LMMs we used can generate high-resolution images, e.g., $1024 \times 1024$, thus Fréchet Inception Distance (FID) is not a perfect metric.
% CLIP-T score measures the similarity between the generated image and the baseline prompt. CLIP-I and DINO scores evaluate the similarity between the generated image and the ground truth one. Also, scores from GPT-4o and BLIP-VQA models offer a reverse evaluation via visual question answering. Additionally, Reward represents the ImageReward score which stands for the human preference.

\paragraph{Large multimodal models} We conducted experiments on two latest LMMs: Emu2-Gen (37B) \cite{DBLP:journals/corr/abs-2312-13286} and Lumina-NEXT (4B) \cite{DBLP:journals/corr/abs-2406-18583}. The former is a large generation model end-to-end trained with 3.8B tokens of language-only data and millions of multimodal interleaved samples. The latter is a large diffusion transformer model integrating with Gemma-2B \cite{DBLP:journals/corr/abs-2403-08295} as its text encoder. We followed their official inference settings and used random seeds.
% Note that for all models, we followed their official default inference settings differing from each other, thus performance comparisons between different models are unfair. We maintained the same generation setups within one model to compare the performance of different ICL methods. All experiments were conducted on NVIDIA A800 GPUs.

\subsection{Quantitative Results}
We compare the performance of PMT2I with that of the original English prompt (EN Prompt) from benchmarks and the automatically optimized prompt (Auto-opt), where Promptist performs the optimization \cite{DBLP:conf/nips/HaoC0W23}. Experimental results are shown in Table \ref{t1}. First of all, we see that PMT2I achieves the best result among all the systems. PMT2I outperforms the established baseline in terms of general T2I synthesis and excels in more complex compositional scenarios and fine-grained assessments. Moreover, we observe significant improvements in the ImageReward scores achieved by PMT2I, indicating that it facilitates the generation of images that align more closely with human preferences.

On the other hand, we apply the reranking method to EN Prompt and PMT2I. All experiments are conducted on 6 candidates except Emu2-Gen+MS-COCO, which involves 3 candidates. We can see that when armed with the reranking method, PMT2I further achieves superior improvements than EN Prompt+rerank, significantly outperforming baselines. Besides, Fig. \ref{f3} (a) shows the L1 distance and reward score of 2000 randomly selected samples. PMT2I samples are distributed at the top-right compared to the ones of EN Prompt, demonstrating PMT2I's superiority in both quality and diversity.

\subsection{Qualitative Results}
Next, we showcase several generated images in Fig. \ref{f2}. We can see that on the one hand, PMT2I improves the factual consistency of the generated images. For instance, in the second sample, EN Prompt confuses the model with the green forest and the green truck. However, when integrated with PMT2I, the model produces a factually accurate image of a green truck within a green forest. On the other hand, PMT2I facilitates text-aligned image generation. In the fourth example, the result of PMT2I fully aligns with the text, including both cat and dog, whereas EN Prompt fails to follow the text by only generating dogs. Indeed, PMT2I is not without its imperfections. We observe that it may introduce duplication errors in rare cases. For example, the last prompt specifies a single bear, yet PMT2I generates a primary bear with a small one at the bottom-left. We hypothesize that the model might individually process the multilingual image descriptions.
% This problem can be solved by using another random seed and generating again.

\subsection{Ablation Study}
We present extensive ablation studies of PMT2I on (1) continually increasing candidates for reranking, (2) dissecting the learning pattern, and (3) monolingual settings. Experiments are conducted on Lumina-NEXT with 10K samples randomly selected from the MS-COCO validation set.

\textbf{$Q_1$: If the PMT2I prompt outperforms the baseline when the number of candidates increases? $A_1$: Yes.} Fig. \ref{f3} (b) shows that as the number of candidates continually grows, PMT2I consistently outperforms the baseline with more significant improvements.

\textbf{$Q_2$: Does the benefits of PMT2I mainly come from a specific language? $A_2$: No.} From Fig. \ref{f3} (c), we can see that English prompts significantly outperform others in non-English languages. Intriguingly, PMT2I leverages multiple languages in conjunction with English to construct prompts that surpass the strong baseline set by English-only prompts, mirroring the \textit{weak-to-strong} learning pattern \cite{DBLP:journals/corr/abs-2402-02416,DBLP:conf/icml/BurnsIKBGACEJLS24}.

\textbf{$Q_3$: If multilingualism is essential for PMT2I? $A_3$: Yes.} To ablate the multilingualism, we compare the performance of the PMT2I prompt with the reduplication and the paraphrasing ones. The former is comprised of $n$ duplicates of the original prompt. The latter consists of $n$ paraphrased texts by GPT-4o. Their templates $f(\cdot)$ remain the same as the PMT2I prompt. Experimental results shown in Fig. \ref{f3} (d) demonstrate the effectiveness of constructing multilingual prompts rather than monolingual ones.

% #################################################
% #################################################
% ################  Conclusion  ###################
% #################################################
% #################################################
\section{Conclusion}
In this work, we propose an ICL method, PMT2I, which triggers the multilingual capabilities of LMMs to boost their performance on T2I generation tasks. PMT2I serves as a simple approach that merely supplies LMMs with the original image description and its machine-translated equivalents in multiple languages. Besides, PMT2I can be scaled to plenty of different prompts, facilitating diverse image synthesis. Extensive experiments demonstrate that PMT2I outperforms vanilla monolingual prompts on general text-image alignment and more granular evaluations, with significant improvements in human preference alignment. Overall, our work not only reveals the similarity of leveraging multilingualism in both LLMs and LMMs but also serves as the first attempt at applying multilingual prompting methods to LMMs on multimodal tasks as far as we know. In future work, we aim to explore additional multilingual ICL methods that benefit LLMs and LMMs.

\newpage

\bibliographystyle{IEEEtran}
\bibliography{refs}

\end{document}